\pdfoutput=1
\documentclass[sigconf, nonacm, screen, authorversion]{acmart}

\let\digamma\relax

\usepackage{graphicx}
\usepackage{booktabs}
\usepackage{amsmath}
\usepackage{pifont}

\begin{document}
	
	\title{ClickRemoval: An Interactive Open-Source Tool for Object Removal in Diffusion Models}
	
	\author{Ledun Zhang}
	\email{202310201057@imut.edu.cn}
	\affiliation{
		\institution{Inner Mongolia University of Technology}
		\city{Hohhot}
		\state{Inner Mongolia}
		\country{China}
	}
	
	\author{Yatu Ji}
	\email{MLjyt@imut.edu.cn}
	\affiliation{
		\institution{Inner Mongolia University of Technology}
		\city{Hohhot}
		\state{Inner Mongolia}
		\country{China}
	}
	
	\author{Xufei Zhuang}
	\email{zxf@imut.edu.cn}
	\affiliation{
		\institution{Inner Mongolia University of Technology}
		\city{Hohhot}
		\state{Inner Mongolia}
		\country{China}
	}
	
	\author{Xinying Yao}
	\email{202310201024@imut.edu.cn}
	\affiliation{
		\institution{Inner Mongolia University of Technology}
		\city{Hohhot}
		\state{Inner Mongolia}
		\country{China}
	}
	\renewcommand{\shortauthors}{Zhang et al.}
	
	\begin{abstract}
	Existing object removal tools often rely on manual masks or text prompts, making precise removal difficult for non-expert users in complex scenes and often leading to incomplete removal or unnatural background completion. To address this issue, we present \textbf{ClickRemoval}, an open-source interactive object removal tool built on pretrained Stable Diffusion models and driven solely by user clicks. Without additional training, hand-drawn masks, or text descriptions, ClickRemoval localizes target objects and restores the background through self-attention modulation during denoising. Experiments show that ClickRemoval achieves competitive results across quantitative metrics and user studies. We release a complete software package at \url{https://github.com/zld-make/ClickRemoval} under the Apache-2.0 license.
	\end{abstract}
	
	\keywords{object removal, click interaction, diffusion models, self-attention control}
	
	\maketitle
	
	\section{Introduction}

	Removing unwanted objects from photos is a common requirement in multimedia content creation, and is widely used in scenarios such as photo editing and privacy protection~\cite{yu2018generative, rombach2022high}. Recently, image generative models have advanced object removal, and representative methods such as AttentiveEraser~\cite{sun2025attentive}, PowerPaint~\cite{zhuang2024task}, and the SD-Inpaint series~\cite{rombach2022high} have achieved significant progress. However, these methods still typically rely on fine-grained manual masks, text prompts, or specialized training pipelines, resulting in relatively high interaction costs and limited usability for non-expert users.

	In this paper, we propose \textbf{ClickRemoval}, an open-source object removal tool based on an attention redirection framework and relying only on click interaction. Users only need to click on the target object, and ClickRemoval automatically performs object localization and background restoration. We provide three implementation configurations: a lightweight version for real-time interaction (SD1.5), a balanced version for general scenarios (SD2.1), and a high-quality version for high-resolution settings (SDXL1.0). Experimental results show that, without requiring additional training, ClickRemoval achieves competitive restoration quality against strong baselines and receives strong user preference across resolutions. In terms of usability, ClickRemoval provides a lightweight point-and-remove workflow, where users only indicate the target object through clicks and the system automatically performs localization, removal, and visually coherent background restoration. In summary, the contributions of this paper are as follows:

	1. \textbf{Interactive object removal tool}: We introduce ClickRemoval, an object removal tool that requires no masks, text descriptions, or additional training, supports positive and negative click interaction, and lowers the usage barrier for non-expert users to achieve precise object removal.

	2. \textbf{Extensible attention redirection mechanism}: We design a click-driven object removal framework composed of the M2N2 semantic distance map, SGAR, SGAS, and ARG. The framework is not tightly coupled to a specific diffusion backbone and can be implemented across SD1.5, SD2.1, and SDXL1.0.

	3. \textbf{Complete open-source delivery}: We release a complete software package, including source code, configuration files, Docker environment, documentation, demo interface, and evaluation scripts. The underlying Stable Diffusion checkpoints are loaded from their official public sources following their original licenses.

	\section{ClickRemoval: Design and Implementation}
		\begin{figure*}[t] 
	\centering 
	\includegraphics[width=0.7\textwidth]{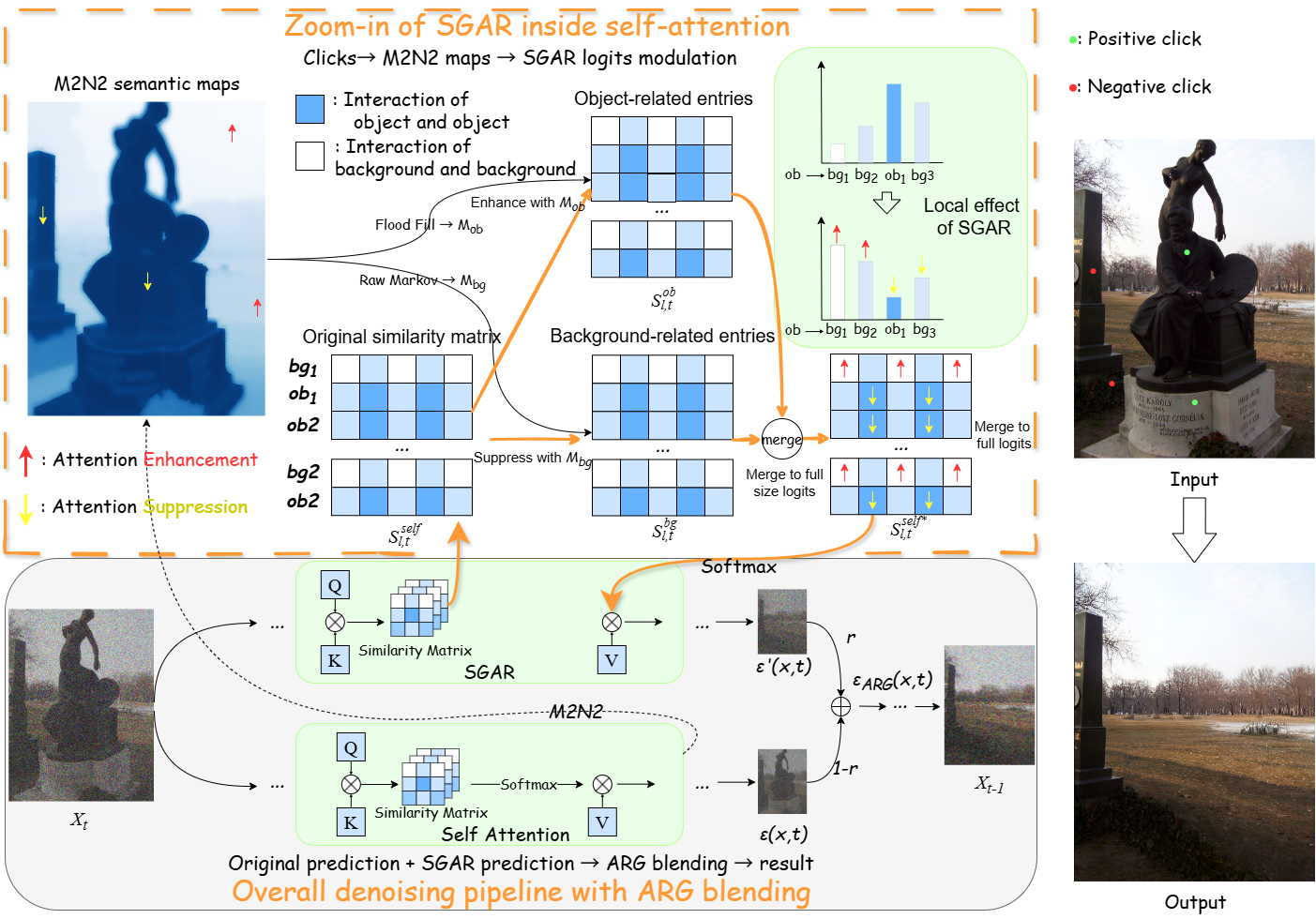} 
	\caption{Overview of ClickRemoval. M2N2 converts user clicks into semantic maps, SGAR and SGAS redirect self-attention during denoising, and ARG blends the original and modulated predictions to control removal strength.}
	\label{fig1} 
\end{figure*}
	Figure~\ref{fig1} illustrates the core mechanism of ClickRemoval. Instead of treating user clicks as hard inpainting masks, ClickRemoval converts them into soft semantic maps and uses these maps to redirect self-attention inside a pretrained Stable Diffusion model. Specifically, M2N2~\cite{karmann2025repurposing} is used as the default click-to-map module to produce a target-related semantic distance map, from which we derive an object map $M_{ob}$ for target suppression and a complementary background-reference map $M_{bg}$ that downweights regions semantically similar to the clicked object. SGAR then modulates object and background related attention logits, SGAS schedules this modulation across denoising steps, and ARG controls the final removal strength by combining the original and modulated noise predictions. The localization and generation guidance are both derived from the Stable Diffusion backbone.

	\subsection{M2N2 Semantic Distance Map Extraction}

	To convert user clicks into semantic distance maps, we follow M2N2 and use self-attention maps from a frozen Stable Diffusion model to construct semantic propagation relations. Following M2N2, this click-to-map stage uses only a single denoising forward pass to collect and aggregate the selected self-attention tensors. Specifically, selected multi-head self-attention maps are aggregated into a transition matrix $A \in \mathbb{R}^{N \times N}$. Starting from the clicked position represented by a one-hot distribution $p_0 \in \mathbb{R}^{1 \times N}$, we perform Markov propagation as $p_n=p_0 \cdot A^n$, where $n \in \{0,1,\dots,n_{\max}\}$. The semantic distance of each position is defined by the minimum propagation step required to reach a relative probability threshold $\tau$.

	For object localization, we apply Flood Fill~\cite{karmann2025repurposing} to suppress local minima and enhance instance awareness, producing the object region $M_{ob}$. For background guidance, we normalize the Markov map without Flood Fill as $M_{bg}$ to measure semantic distance from the clicked object, where smaller values indicate stronger semantic similarity to the object.

	\subsection{Self-Guided Attention Redirection and Scheduling}
	\begin{table*}[t]
	\centering
	\setlength{\tabcolsep}{4pt}
	\normalsize
	\caption{Quantitative comparison with state-of-the-art image inpainting and object removal methods (the upper half corresponds to 512×512 resolution inference, and the lower half corresponds to 1024×1024 resolution inference).}
	\footnotesize
	\begin{tabular}{lccccccccc}
		\toprule
		Method   & Interaction    & Extra Training    & FID$\downarrow$   & KID($\times10^{-3}$)$\downarrow$ & Local-FID$\downarrow$  & Inference Time & Memory & User Study & GPT Validation \\
		\midrule
		SD1.5-Inp\textsuperscript{\textsl{CVPR'22}}      & M  & \ding{51}   & 27.63  & 13.53 & 32.23 &2.97s &9.37G&14.64\%&11.53\%\\
		SD1.5-Inp\textsuperscript{\textsl{CVPR'22}}      & M+T  & \ding{51}   & 27.17  & 12.70 & 31.65 &2.97s &9.37G&10.98\%&3.85\%\\
		LaMa\textsuperscript{\textsl{CVPR'22}} & M  & \ding{51}   & 21.63  & 11.70 & 24.10 &0.82s &1.11G&13.41\%&23.08\%\\
		PixelHacker\textsuperscript{\textsl{arXiv'25}} & M  & \ding{51}   & 26.01  & 13.71 & 28.40 &1.77s&9.49G&20.73\%&23.08\% \\
		InpaintAnything\textsuperscript{\textsl{arXiv'23}} &  C  & \ding{51}  & 23.55& 12.15 & 25.29 & 8.20s &6.86G&9.75\%&9.61\%\\	
		Ours (SD1.5) & C & \ding{55}  & 9.35  & 0.899 & 17.27 &4.75s &8.38G&30.48\%&28.85\%\\
		\midrule
		BrushNet\textsuperscript{\textsl{ECCV'24}} & M+T  & \ding{51}  & 25.34  & 6.64 & 42.80 & 21.46s &8.75G&0\%&0\%\\
		PowerPaint-v2\textsuperscript{\textsl{ECCV'24}} & M  &\ding{51}   & 11.11  & 1.45 & 21.91 &22.08s &5.54G&18.92\%&19.60\%\\
		AttentiveEraser\textsuperscript{\textsl{AAAI'25}} & M  & \ding{55}  & 7.98  & 0.557 & 15.74 &21.00s &9.90G&39.18\%&41.18\%\\		
		Ours(SDXL1.0) & C & \ding{55}  & 8.05  & 0.556 & 15.56 &19.86s &15.32G&41.89\%&39.21\%\\
		
		\bottomrule
		\multicolumn{10}{l}{\scriptsize M denotes Mask, T denotes Text, and C denotes Click. Results are compared within the same resolution group.} 
	\end{tabular}
	\label{tab1}
\end{table*}
\begin{figure*}[t]
	\centering
	\includegraphics[width=0.95\textwidth]{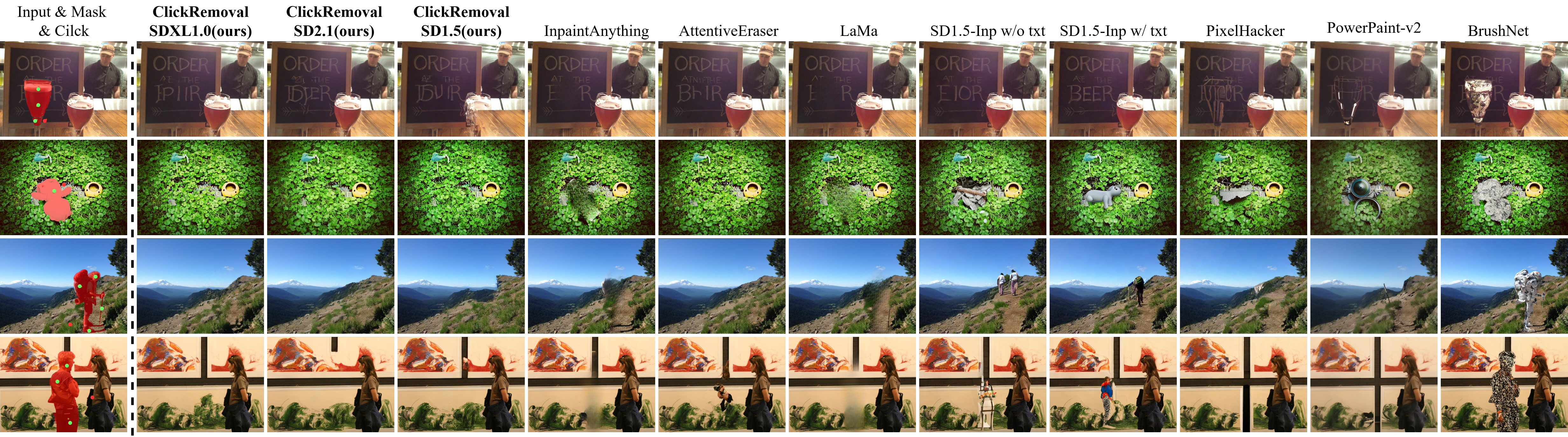}
	\caption{Qualitative comparison with baseline methods. Green points indicate positive clicks for removal, and red points indicate negative clicks for preservation.} 
	\label{fig2} 
\end{figure*}
	To accurately suppress the target object and naturally restore the background, ClickRemoval modulates the self-attention of a pretrained Stable Diffusion model during inference. This process consists of two collaborative components: SGAR modifies the attention distribution within each denoising step, while SGAS controls the strength and timing of SGAR throughout the denoising process.
	
	\textbf{SGAR.} Let $S_{\mathrm{self}} \in \mathbb{R}^{N \times N}$ be the self-attention logits before softmax, and let $M_{ob}$ and $M_{bg}$ denote the object map and background semantic distance map. SGAR redirects attention by suppressing object-related key entries and reweighting background key entries. Specifically, we use $S'_{\mathrm{self}}=S_{\mathrm{self}}\odot\mathcal{B}(G_{bg}(t))+\mathcal{B}(P_{ob})$, where $P_{ob}$ is a large negative object-key penalty, $G_{bg}(t)=1-(1-\alpha(t))\widetilde{M}_{bg}$ is the scheduled background reweighting factor, and $\mathcal{B}(\cdot)$ broadcasts spatial maps to query-key entries. Here $\widetilde{M}_{bg}$ is derived from $M_{bg}$, and $\alpha(t)$ controls the SGAS schedule. We apply SGAR to decoder self-attention layers after resizing the maps to the corresponding attention resolution.
	
	\textbf{SGAS.} Applying strong guidance throughout the entire denoising process may degrade the naturalness of background completion. Therefore, we introduce SGAS as a staged scheduling strategy. Specifically, the early denoising stage, usually the first $20\%$ of steps, is left unchanged. In the early-middle stage, SGAR is enabled, and background guidance gradually decays according to $\alpha(t)$. In the middle stage, background guidance is disabled while object suppression is retained. In the late stage, all guidance is disabled, allowing the model to complete the remaining details by itself. In practice, background guidance is usually needed only for a few early-middle steps, typically about 5--10 steps.

	\subsection{Attention Redirection Guidance}
	
	Although SGAS combined with SGAR can suppress the target object and guide the model toward the background, it produces a fixed modulated noise prediction $\epsilon'(\mathbf{x}, t)$. To provide controllable output-level guidance, we further introduce ARG, inspired by ERG~\cite{ifriqi2025entropy}. Instead of directly modifying self-attention, ARG linearly combines the original and modulated noise predictions as~$\epsilon_{\mathrm{ARG}}=(1-r)\cdot \epsilon(\mathbf{x},t)+r \cdot \epsilon'(\mathbf{x},t)$, where $r$ is a user-specified guidance strength coefficient. A smaller $r$ keeps the result closer to the original diffusion model, while a larger $r$ makes it closer to the SGAR-modulated prediction.

	\section{Experiments}
	
	\subsection{Experimental Setup}

	Recent object removal studies have pointed out that many methods still follow the image inpainting evaluation protocol, using the original image with the target object as the reference for metrics such as Fr\'echet Inception Distance (FID)~\cite{suvorov2022resolution, sun2025attentive}. This setting is unsuitable for object removal, since the goal is to remove the specified target and naturally restore the background, while the original image still contains the object~\cite{oh2024object,fathi2025aura,chandrasekar2024remove}. We therefore construct a unified object removal test set based on Pico-Banana-400K~\cite{qian2025pico}, filtering object-removal samples and using the edited target-free images as references. Due to the license restrictions of Pico-Banana-400K, we do not redistribute the original, edited, or derived benchmark images, but release the evaluation protocol, annotation schema, and scripts for users with official dataset access. After manual cleaning and annotation, we obtain approximately 5{,}000 test samples covering diverse scenes, object categories, and mask sizes, with both clicks and masks manually annotated.
	
	We compare ClickRemoval with open-source inpainting and object removal baselines, including Stable Diffusion 1.5 Inpainting under mask-only and mask-plus-text settings~\cite{rombach2022high}, LaMa~\cite{suvorov2022resolution}, AttentiveEraser~\cite{sun2025attentive}, PixelHacker~\cite{xu2025pixelhacker}, click-guided Inpaint Anything with the Remove Anything pipeline~\cite{yu2023inpaint}, PowerPaint-v2~\cite{zhuang2024task}, and BrushNet~\cite{ju2024brushnet}. All baselines use their official inference configurations and public pretrained weights. We report FID, Kernel Inception Distance (KID), and Local-FID~\cite{xie2023smartbrush} for global and local restoration quality. For Local-FID, we crop a square region centered at the mask bounding box, with side length $L=\max(L_{bbox},299)$, where $L_{bbox}$ is the longer side of the bounding box.
	
	\subsection{Comparison Experiments and Restoration Effect Visualization}

	As shown in Table~\ref{tab1}, ClickRemoval achieves competitive restoration quality without additional training. At 1024 resolution, the SDXL1.0 variant of ClickRemoval reaches FID 8.05 and Local-FID 15.56, comparable to AttentiveEraser (7.98 / 15.74). At 512 resolution, the SD1.5 variant of ClickRemoval obtains the best overall performance among same-resolution methods, with FID 9.35, KID 0.899, and Local-FID 17.27. We further evaluate runtime and GPU memory on an RTX 3090 24G consumer GPU. ClickRemoval shows acceptable inference overhead, with 1024-resolution runtime comparable to strong baselines and practical 512-resolution efficiency for click-driven restoration.
	
	Figure~\ref{fig2} shows representative results on text, grass, landscape, and human scenarios. ClickRemoval removes target objects more completely while producing more natural background textures and structures.
	
\begin{figure}[t]
	\centering
	\includegraphics[width=0.4\textwidth]{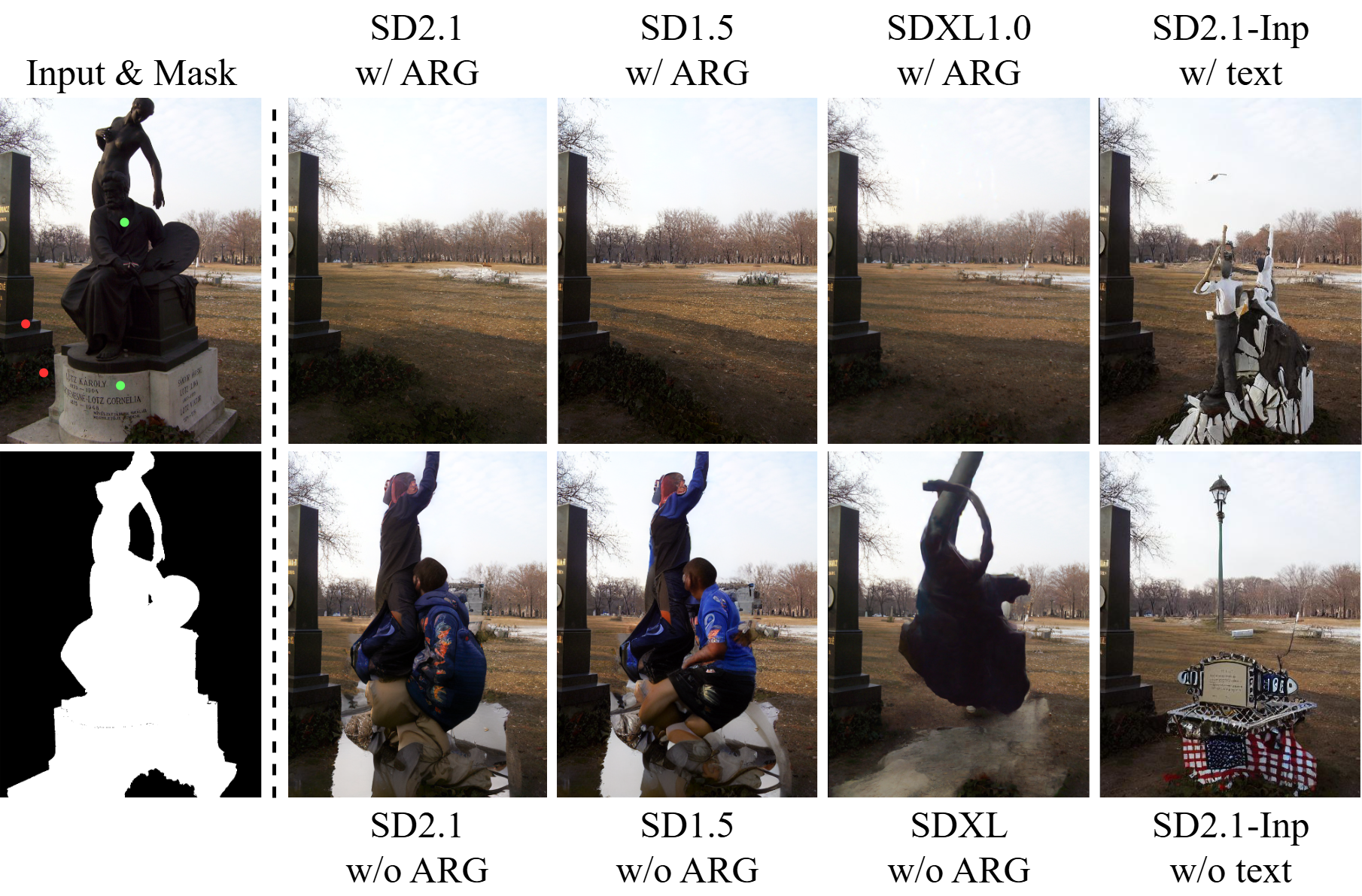}
	\caption{Ablation comparison of different model variants on challenging removal cases.}
	\label{fig3}
\end{figure}

	To investigate the effects of different backbone variants and text prompts on restoration performance, we conduct ablation studies, as shown in Fig.~\ref{fig3}. The results show that vanilla SD backbones, although not originally designed for object removal, can successfully perform target removal and background restoration when equipped with our proposed ARG. In contrast, SD1.5-Inp struggles to remove the target object thoroughly, regardless of whether text prompts are provided.
	
\begin{figure}[t]
		\centering
		\includegraphics[width=0.4\textwidth]{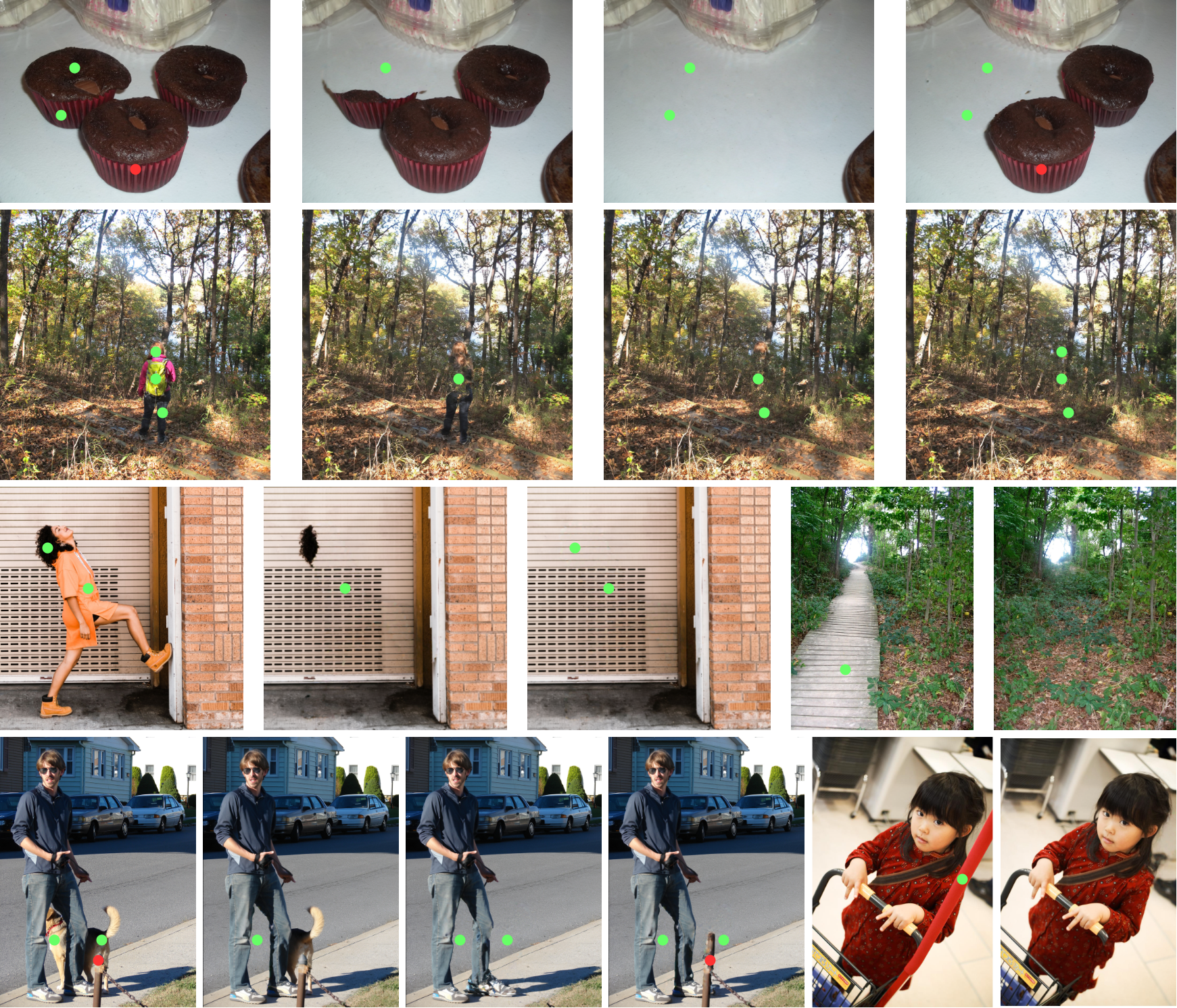}
		\caption{Progressive editing results with additional positive and negative clicks.} 
		\label{fig4}
	\end{figure}

	To evaluate the interactive capability of ClickRemoval, Fig.~\ref{fig4} shows progressive restoration results under different click settings, including large-object removal, occluded-object removal, multi-object removal, and complex-background restoration. We compare results using a single positive click, two positive clicks, and multiple positive and negative clicks. More positive clicks generally improve removal completeness, while negative clicks help preserve non-target regions. For example, negative clicks prevent the pillar from being removed when it occludes the target dog, and allow only the specified cake to be removed in a multi-cake scene.

	\subsection{User Study and LLM-Assisted Validation}

	To complement the limitations of quantitative metrics for object removal, we conduct both a user preference study and GPT-assisted validation. For both evaluations, we randomly select 50 images from the test set and evaluate the results separately at 512 and 1024 resolutions to avoid bias caused by resolution differences. The user preference study involves 25 non-expert participants, who are asked to select the result with the best overall visual quality and removal effectiveness. For GPT-assisted validation, we use a GPT-based multimodal evaluator with the same evaluation prompt for all samples. The evaluator is given the original image, the mask, and the restoration results of different methods with method names hidden, and is asked to select the result with the best overall object removal and background restoration quality. The results are reported in Table~\ref{tab1}.
	
	In the user preference study, ClickRemoval receives 30.48\% of the votes at 512 resolution and 41.89\% at 1024 resolution, ranking first in both settings. In GPT-assisted validation, ClickRemoval also ranks first at 512 resolution with a score of 28.85\%. At 1024 resolution, it obtains 39.21\%, ranking second and only 1.97 percentage points behind AttentiveEraser (41.18\%). Overall, these results show that ClickRemoval achieves competitive restoration quality against strong baselines without requiring additional training.
	
	\section{Conclusion}
	We present ClickRemoval, a training-free open-source tool for interactive object removal using only user clicks. By combining click-based semantic maps, scheduled self-attention redirection, and adaptive restoration guidance, ClickRemoval removes target objects without manual masks or text prompts. Experiments show competitive restoration quality and strong user preference across resolutions. The released code, model download scripts, Docker environment, and documentation are intended to support reproducible research and practical content editing.

	\bibliography{references}
	
\end{document}